\documentclass[11pt]{article}
\usepackage{acl2016}
\usepackage{times}
\usepackage{url}
\usepackage{latexsym}
\usepackage{graphicx,multirow}
\usepackage{tabularx}
\usepackage{lineno}
\usepackage{booktabs}
\newcolumntype{Y}{>{\centering\arraybackslash}X}
\usepackage{placeins}

\makeatletter
\newcommand{\@BIBLABEL}{\@emptybiblabel}
\newcommand{\@emptybiblabel}[1]{}
\makeatother
\usepackage[draft, hidelinks]{hyperref}

\aclfinalcopy % Uncomment this line for the final submission
\title{Shallow Discourse Parsing Using Distributed Argument Representations and Bayesian Optimization}

\author{Akanksha \\
  Georgia Institute of Technology  \\
  {\tt akanksha271@gmail.com} \\\And
  Jacob Eisenstein \\
  Georgia Institute of Technology \\
  {\tt jacobe@gmail.com} \\}

\date{}

\begin{document}
\maketitle
\begin{abstract}
This paper describes the Georgia Tech team's approach to the CoNLL-2016 supplementary evaluation on discourse relation sense classification. We use long short-term memories (LSTM) to induce distributed representations of each argument, and then combine these representations with surface features in a neural network. The architecture of the neural network is determined by Bayesian hyperparameter search.
\end{abstract}

\section{Introduction}
Our approach to discourse relation classification is to combine strong surface features with a distributed representation of each discourse unit. This follows prior work demonstrating that distributed representations can improve generalization for this task~\cite{ji2014representation,ji2015one,braud2015comparing}. We combine these two disparate representations in a neural network architecture. Our approach is shaped by two main design decisions: the use of long short-term memory recurrent networks~\cite{hochreiter1997long} to induce representations of each discourse unit, and the use of Bayesian optimization~\cite{snoek-etal-2012b} for tuning the neural network architecture.

\section{System Overview}
\begin{figure}
\includegraphics[width=8cm]{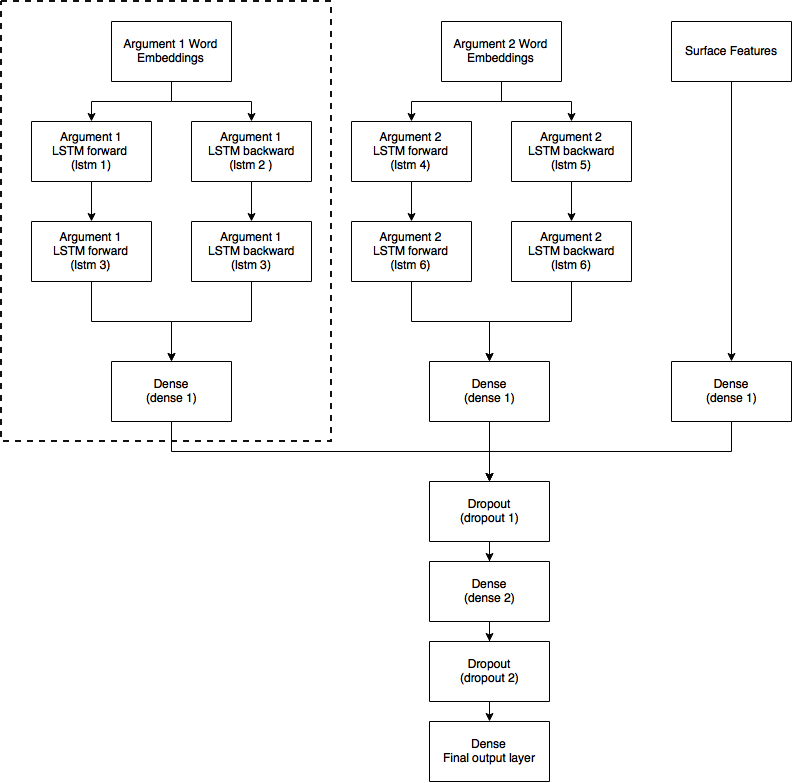}
\caption{System Architecture}
\label{fig:arch}
\end{figure}

The overall architecture is shown in \autoref{fig:arch}. The same architecture is used for both explicit and non-explicit relations, but with different parameters. The output of the classifier is a softmax layer, which takes as input a series of dense layers. These dense layers allow nonlinear interactions between surface features and elements of the distributed representations. Dropout is employed to reduce overfitting~\cite{srivastava2014dropout}. The overall architecture is trained to minimize cross-entropy. The implementation is in Keras~\cite{chollet2015keras}, and training takes several hours on a standard CPU. We now describe of the subcomponents of the classifier in detail.

\subsection{Distributed representations for discourse units} 
First, we induce representations for each unit in each discourse relation. This component of the model is shown in the dotted part of \autoref{fig:arch} for the first discourse argument. Prior work has explored a variety of ways for inducing representations of discourse units, including average pooling~\cite{ji2014representation,braud2015comparing} and recursive neural networks on syntactic parse trees~\cite{li2014recursive,ji2015one}. We take a recurrent neural network approach, characterizing each discourse unit by a recurrently-updated state vector~\cite{li2015when}, with the input consisting of pre-trained word embeddings \texttt{GoogleNews-vectors-negative300.bin} from the \texttt{word2vec} page.\footnote{https://code.google.com/archive/p/word2vec/}

Specifically, our recurrent architecture is a long short-term memory (LSTM), which uses a combination of gates to better handle long-term dependencies, as compared with the more straightforward recurrent neural network~\cite{hochreiter1997long}. Following \newcite{graves2005framewise}, we employ a bidirectional LSTM, in which each training sequence is presented forwards and backwards to two separate recurrent nets, both of which are connected to the same output layer. We combine the output of these bidirectional LSTMs in a multilayer perceptron with the extracted surface features.

\subsection{Surface features}
In addition to the distributed representations of the discourse units, we use some of the most successful surface features from prior work. These features are implemented using the Natural Language Toolkit~\cite{bird2009natural} and scikit-learn~\cite{pedregosa2011scikit}. In general, these features were inspired by the system from~\newcite{wang2015refined}, which obtained best performance on the PDTB test set in the 2015 shared task~\cite{xue2015conll}.

\subsubsection{Features for explicit relations}
\begin{description}
\item[Connective Text] The connective itself is a strong feature for sense classification of explicit discourse relations~\cite{pitler2008easily}. This feature alone yields F1 of 0.8862 for our classifier.
\item[Sentiment Value] The Vader Sentiment analysis package~\cite{hutto2014vader} was used to calculate sentiment score for both arguments. The feature then reports whether the two arguments have the same sentiment.
\item[Trigrams] We used trigram features for the final three words of arg1, and for the first three words of arg2.
\end{description}

\begin{table}[!t]
\center
\tabcolsep 3pt
\begin{tabular}{@{}llcc@{}}
\toprule
\textbf{Hyperparameter} & \textbf{Range} & \textbf{Best} \\
\midrule
\emph{Number of hidden nodes} &&\\
\multirow{1}{*}{lstm1 } &64-320& $259$  \\
\multirow{1}{*}{lstm2 } &64-100& $75$  \\
\multirow{1}{*}{lstm3 } &64-320& $263$  \\
\multirow{1}{*}{lstm4}  &64-320& $127$  \\
\multirow{1}{*}{lstm5}  &64-100& $89$  \\
\multirow{1}{*}{lstm6}  &64-320& $150$  \\
\multirow{1}{*}{dense1} &64-320& $269$  \\
\multirow{1}{*}{dense2} &64-100& $69$  \\[5pt]
\emph{Percentage of dropout} &&\\
\multirow{1}{*}{dropout1} &0-0.9& $0.11$  \\
\multirow{1}{*}{dropout2} &0-0.9& $0.57$  \\[5pt]
\emph{Learning Rate} &&\\
\multirow{1}{*}{SGD} &0.001-0.5& $0.1549$  \\
\bottomrule
\end{tabular}
\caption{Hyperparameters selected by Spearmint from the provided ranges, for non-explicit discourse relations}
\label{tab:best-params}
\end{table}

\subsubsection{Features for non-explicit relations}
We used the same \textbf{trigrams} features from the explicit relation classifier, as well as the following additional features on pairs of linguistic elements in arg1 and arg2. 
\begin{description}
\item[Word Pairs] We formed word pairs from the cross product of all words appearing in arg1 and arg2, following much of the prior work in discourse parsing~\cite{marcu2002unsupervised,pitler2009automatic}. We then replaced the words in each pair with a cluster identity~\cite{rutherford2014discovering}. Specifically, we used the GoogleNews-vectors-negative skipgram word embeddings to form 1000 clusters.
\item[Part-of-Speech Pairs] Similarly, we formed part-of-speech pairs from the tags appearing in the two arguments~\cite{rutherford2014discovering}.
\item[Production Rules Pairs] Using the syntactic analysis of each argument, we form pairs of production rules appearing in the two arguments~\cite{lin2009recognizing}.
\item[Adverb Pairs] Adverbs are particularly relevant for non-explicit discourse relations, so we compute features from pairs of adverbs appearing the two arguments.
\end{description}

\begin{figure}
  \centering
  \includegraphics[width=0.5\textwidth]{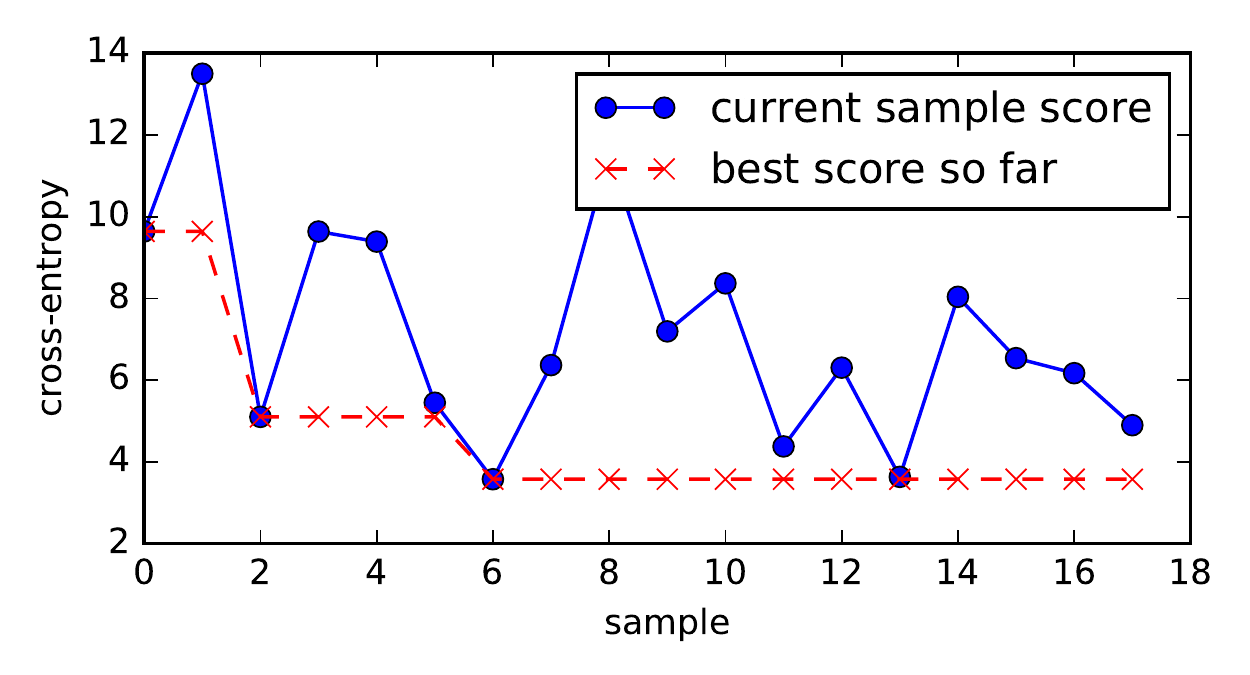}
  \caption{Progress of Bayesian optimization over hyperparameter space}
  \label{fig:opt-path}
\end{figure}

\subsection{Hyperparameter tuning}
The best set of hyperparameters for the classifiers were found using spearmint~\cite{snoek-etal-2012b}, using the \texttt{GPEIOptChooser} Markov Chain Monte Carlo search algorithm. This algorithm samples from the space of hyperparameters, while trying to learn a function from hyperparameters to overall performance. We use the cross-entropy (and not the F1) as the overall performance objective; due to the time cost for training the model, we could run only twenty samples, which took several days. The progress of this search is shown in \autoref{fig:opt-path}, which indicates that the best hyperparameter configuration was identified on the sixth sample. More samples may have yielded better performance, but this was not possible under the time constraints of the shared task. The best set of hyperparameters for non-explicit discourse relation classification are listed in \autoref{tab:best-params}. 
\begin{table}
\center
\tabcolsep 3pt
\begin{tabular}{@{}llcc@{}}
\toprule
\addlinespace
&\textbf{Feature Type}& \textbf{$F1$} \\
\midrule
\multirow{10}{*}{Non-Explicit} & Distributed & $0.3485$ \\
&  + Argument 2 first 3 & $0.3872$ \\
&  + Argument 1 last 3 & $0.3044$ \\
&  + Word Pairs& $0.3672$ \\
&  + Parts of Speech & $0.3672$ \\
&  + Adverbs& $0.3979$ \\
&  + Inquirer& $0.3979$ \\
&  + Production Rule& $0.4072$ \\
\midrule
\multirow{3}{*}{Explicit} & Distributed & $0.3839$ \\
& + Connective & $0.8862$  \\
& + Sentiment & $0.9029$\\
& + Argument 2 first 3 & $0.8816$ \\
& + Argument 1 last 3& $0.8983$\\
\bottomrule
\end{tabular}
\caption{Evaluation as the features are added incrementally to the purely distributed model.}
\label{tab:feature-results}
\end{table}

%%%%%%%%%%%%%%%%%%%%%%%%%%%%%%%%%%%%%%%%%%%%%%%%%%%%%%

\section{Evaluation}

\begin{table*}
  \centering
  \small
  \begin{tabular}{llllllllll}
    \toprule
    & \multicolumn{3}{c}{Dev}
    & \multicolumn{3}{c}{Test}
    & \multicolumn{3}{c}{Blind} \\
    System & All & Explicit & Non-expl.
           & All & Explicit & Non-expl.
           & All & Explicit & Non-expl. \\
    \midrule
    \texttt{tbmihaylov} 
    & 0.641 & 0.912 & 0.403
    & 0.633 & 0.898 & 0.392
    & \textbf{0.546} & \textbf{0.782} & 0.345 \\
    \texttt{ecnuc} & \textbf{0.680} & \textbf{0.926} & \textbf{0.464}           & \textbf{0.643} & \textbf{0.901} & \textbf{0.409}
    & 0.541 & 0.774 & 0.342 \\
    \texttt{gtnlp} (this paper) & 0.639
 & 0.903 & 0.407
& 0.609 & 0.895 & 0.350
    & 0.543 & 0.750 & \textbf{0.368}\\
    \bottomrule
  \end{tabular}
  \caption{Discourse sense classification results, measured by F1, in comparison with the most competitive systems from the shared task.}
  \label{tab:overall-results}
\end{table*}

\iffalse
\begin{table}
\begin{minipage}{\textwidth/2}
\tabcolsep 4pt
\begin{tabular}{|p{2cm}|c c c|}
\hline
& \multicolumn{3}{c|}{{ { Sense Performance F1}}} \\
System & All & Explicit & Non-Explicit\\
\hline
Baseline&0.2202&0.2807&0.1669\\\hline
\multicolumn{4}{|c|}{{{\normalsize comparison on dev dataset}}}\\\hline
System 01 &0.6797&0.9256&0.4642\\\hline
System 02 &0.6413&0.912&0.4032\\\hline
Our System &0.6392&0.9029&0.4072\\\hline
\multicolumn{4}{|c|}{{{\normalsize comparison on blind test dataset}}}\\\hline
System 01 &0.5406&0.7741&0.3418\\\hline
System 02 &0.546&0.782&0.3451\\\hline
Our System &0.543&0.7495&0.3675\\\hline
\end{tabular}
\caption[]{%
Result comparison with the baseline on dev dataset and conll16 shared task best systems on the dev dataset (System 01) %
    \footnote{discourse-relation-sense-classification-ecnuc}%
    and blind test dataset(System 02)%
  \footnote{discourse-relation-sense-classification-tbmihaylov}%
}
\label{tab:overall-results}
\end{minipage}
\end{table}
\fi
\begin{table*}[!t]
\center
\small
\em
\begin{tabular}{ccccccc}
\toprule
& \multicolumn{3}{c}{{\em {\normalsize Explicit}}} & \multicolumn{3}{c}{{\em {\normalsize Non-Explicit}}} \\
\addlinespace
& precision & recall & F1 & precision & recall & F1  \\
\cmidrule(lr){2-4} \cmidrule(lr){5-7}
\multirow{1}{*}{{\em {\normalsize Micro-Average}}} & 0.9029& 0.9029	& 0.9029&  0.4072& 0.4072	& 0.4072\\
\addlinespace
{\em {\normalsize Comparison.Concession}} & 1.0000	& 0.0833	& 0.1538 & 1.0000	& 0.0000	& 0.0000\\
\addlinespace
{\em {\normalsize Comparison.Contrast}} & 0.9387	& 0.9563	& 0.9474 & 0.2391	& 0.2619	& 0.2500\\
\addlinespace
{\em {\normalsize Contingency.Cause.Reason}} & 0.8235	& 0.6829	& 0.7467 & 0.3714	& 0.1688	& 0.2321\\
\addlinespace
{\em {\normalsize Contingency.Cause.Result}} & 1.0000	& 0.8421	& 0.9143 &0.3714	& 0.1688	& 0.2321\\
\addlinespace
{\em {\normalsize Contingency.Condition}} & 0.9778	& 0.9362	& 0.9565 & -	& -	& -\\
\addlinespace
{\em {\normalsize EntRel}} & -	& -	& - & 0.5143	& 0.7535	& 0.6113\\
\addlinespace
{\em {\normalsize Expansion.Alternative}} & 0.8571	& 1.0000	& 0.9231 & -	& -	& -\\
\addlinespace
{\em {\normalsize Expansion. Alternative.Chosen alternative}} & 1.0000	& 0.8333	& 0.9091 & 1.0000	& 0.0000	& 0.0000\\
\addlinespace
{\em {\normalsize Expansion.Conjunction}} & 0.9286	& 0.9891	& 0.9579 & 0.3298	& 0.5122	& 0.4013\\
\addlinespace
{\em {\normalsize Expansion.Instantiation}} & 1.0000	& 1.0000	& 1.0000 & 0.4783	& 0.2292	& 0.3099\\
\addlinespace
{\em {\normalsize Expansion.Restatement}} & 0.3750	& 0.2621	& 0.3086 & 0.3750	& 0.2621	& 0.3086\\
\addlinespace
{\em {\normalsize Temporal.Asynchronous.Precedence}} & 0.9608	& 1.0000	& 0.9800 & 0.2857	& 0.0800	& 0.1250\\
\addlinespace
{\em {\normalsize Temporal.Asynchronous.Succession}} & 1.0000	& 0.6667	& 0.8000 & 1.0000	& 0.0000	& 0.0000\\
\addlinespace
{\em {\normalsize Temporal.Synchrony}} & 0.6842	& 0.9420	& 0.7927 & 1.0000	& 0.0000	& 0.0000\\

\bottomrule
\end{tabular}
\caption{Dev set evaluation for explicit and non-explicit (Implicit, EntRel, AltLex) discourse relations}
\label{tab:results-by-relation}
\end{table*}

Evaluation was performed using the evaluation script provided by the conll16 task organizers. In \autoref{tab:overall-results}, the performance of our system is compared to the best-performing systems from this year's shared task. Our system was particularly competitive on the blind test set. (The best performance on non-explicit relations on the blind test set was from \texttt{ttr}, but this system did not attempt to classify explicit relations, and so did not obtain an overall score for all relations.) This suggests that our approach suffered less from overfitting to the dev data. On the other hand, our system's performance on explicit relations was further behind the best systems, suggesting the need for additional features to handle this case.

Results are broken down by individual relation types in \autoref{tab:results-by-relation}, again using the shared task evaluation script. In addition, the contribution of each feature is indicated in \autoref{tab:feature-results}, in which features are incrementally added to a baseline model containing only the distributed representations of each argument.

\paragraph{Acknowledgments} We thank the organizers of the shared task for formalizing this evaluation, and we thank Yangfeng Ji for helpful discussions in the initial phase of the project.

%%%%%%%%%%%%%%%%%%%%%%%%%%%%%%%%%%%%%%%%%%%%%%%%%%%%%%%%%%%%%

\FloatBarrier
\bibliography{cite-strings,cites,cite-definitions,acl2016}
\bibliographystyle{acl2016}

\end{document}